\documentclass[pdflatex,sn-mathphys-num]{sn-jnl}

\usepackage{graphicx}%
\usepackage{multirow}%
\usepackage{amsmath,amssymb,amsfonts}%
\usepackage{amsthm}%
\usepackage{mathrsfs}%
\usepackage[title]{appendix}%
\usepackage{xcolor}%
\usepackage{textcomp}%
\usepackage{manyfoot}%
\usepackage{booktabs}%
\usepackage{algorithm}%
\usepackage{algorithmicx}%
\usepackage{algpseudocode}%
\usepackage{listings}%

\usepackage{booktabs}
\usepackage{pifont}
\usepackage{makecell}
\usepackage{parskip}

\newcommand{\cmark}{\ding{51}}

\theoremstyle{thmstyleone}%

\theoremstyle{thmstyletwo}%

\theoremstyle{thmstylethree}%

\begin{document}

\title[Article Title]{From Observation to Insight: Mechanistic World Models and the Quest for Autonomous Discovery}

\author*[1]{\fnm{Ingmar} \sur{Posner}}\email{ingmar@robots.ox.ac.uk}

\author[1]{\fnm{Anson} \sur{Lei}}\email{anson@robots.ox.ac.uk}

\author[2]{\fnm{Bernhard} \sur{Schölkopf}}\email{bs@tuebingen.mpg.de}

\affil*[1]{\orgdiv{Applied AI Lab.}, \orgname{University of Oxford}, \country{United Kingdom}}

\affil[2]{\orgname{MPI for Intelligent Systems \& ELLIS Institute}, \city{Tübingen}, \country{Germany}}

\abstract{Recent advances in foundation models have transformed AI for Science, enabling remarkably accurate predictive performance across domains ranging from protein folding to weather forecasting. Yet prediction alone does not constitute scientific discovery. Scientific understanding depends on uncovering the reusable explanatory mechanisms that generate observations, whereas contemporary machine learning remains fundamentally organised around predictive mappings rather than explanatory structure. In this paper, we argue that scientific discovery is fundamentally a problem of knowledge organisation. To this end, we introduce \emph{Mechanistic World Models}, a new design paradigm that places reusable mechanisms at the centre of representation, computation and learning. Drawing on insights from the philosophy of science, we derive the computational capabilities required for discovery, identify the design principles and inductive pressures that encourage explanatory knowledge to emerge, and formalise the anatomy of a mechanism-centric world model. Finally, we show how diverse research directions including mechanistic interpretability, causal representation learning, equation discovery and modular architectures capture complementary ingredients of this paradigm while lacking a unified framework. We propose Mechanistic World Models as a conceptual foundation and computational blueprint for moving AI beyond predictive forecasting towards autonomous scientific discovery.
}

\maketitle

\section{Introduction}
Across the sciences, we are drowning in data but starved of insight. Recent breakthroughs in domains such as protein folding
(e.g.~\cite{abramson2024accurate})
and weather forecasting (e.g.~\cite{lam2023graphcast, Price2024GenCast,Bonev2025FourCastNet3}) demonstrate deep learning's predictive power and the enormous impact AI can have on our collective scientific endeavour. Yet these models remain \emph{predictors, not explainers}: they map inputs to outputs but do not reveal the reusable mechanisms that underpin scientific understanding~\cite{terwilliger2024alphafold,Krokidis2025AlphaFold3Overview,Bonavita2024LimitationsMLWeather}. 
Scientific discovery relies on the ability to uncover hidden structure in the world, to look beyond observation and arrive at \emph{insight}. Archimedes did not simply notice that some objects float; he revealed the principle of buoyancy. 
This capacity to look past raw observation allows us to distil complexity into explanatory principles that can be reused across phenomena, providing the fundamental building blocks from which scientific knowledge is constructed.
Current AI systems can undoubtedly accelerate scientific discovery by generating predictions, hypotheses and candidate solutions. Yet the knowledge they acquire remains fundamentally predictive rather than explanatory. 
Their performance scales with data and compute~\cite{kaplan2020scaling,hoffmann2022training}, but remains brittle under distribution shift, often relying on spurious correlations~\cite{damour2020underspecification} rather than the structured abstractions required for robust reasoning~\cite{scholkopf2022causality}. Our central thesis is that today's models provide an effective substrate for \emph{AI for Science}, but are fundamentally ill-suited for \emph{Scientific Discovery}. Scientific discovery is not simply a case of making better predictions, but of acquiring explanatory knowledge.

World Models (WMs), which learn to forecast future states of an environment from observations (e.g.~\cite{HaSchmidhuber2018WorldModels,Hafner2025DreamerV3Nature,parkerholder2024genie2}), provide a particularly compelling foundation for addressing this challenge. By explicitly modelling how systems evolve under observation and intervention, they naturally mirror the scientific cycle of hypothesis, experiment and refinement. However, like today's predictive models more broadly, conventional world models remain organised around forecasting observations rather than discovering the reusable mechanisms that generate them. 

This raises a fundamental question: if prediction alone is insufficient, what computational capabilities are required for scientific discovery? Philosophy of Science (PoS) has spent decades analysing the nature of scientific discovery, explanation and understanding. We therefore turn to this body of work not merely as historical background, but as a source of design principles for future AI systems. In particular, we ask what properties a learned model must possess in order to support scientific discovery.
The answer to this question motivates the central proposal of this paper: \emph{Mechanistic World Models} (MWMs), a new design paradigm that shifts the objective of representation learning from modelling observations to modelling the mechanisms that generate those observations.

\begin{figure}
    \centering
    \includegraphics[width=0.99\linewidth]{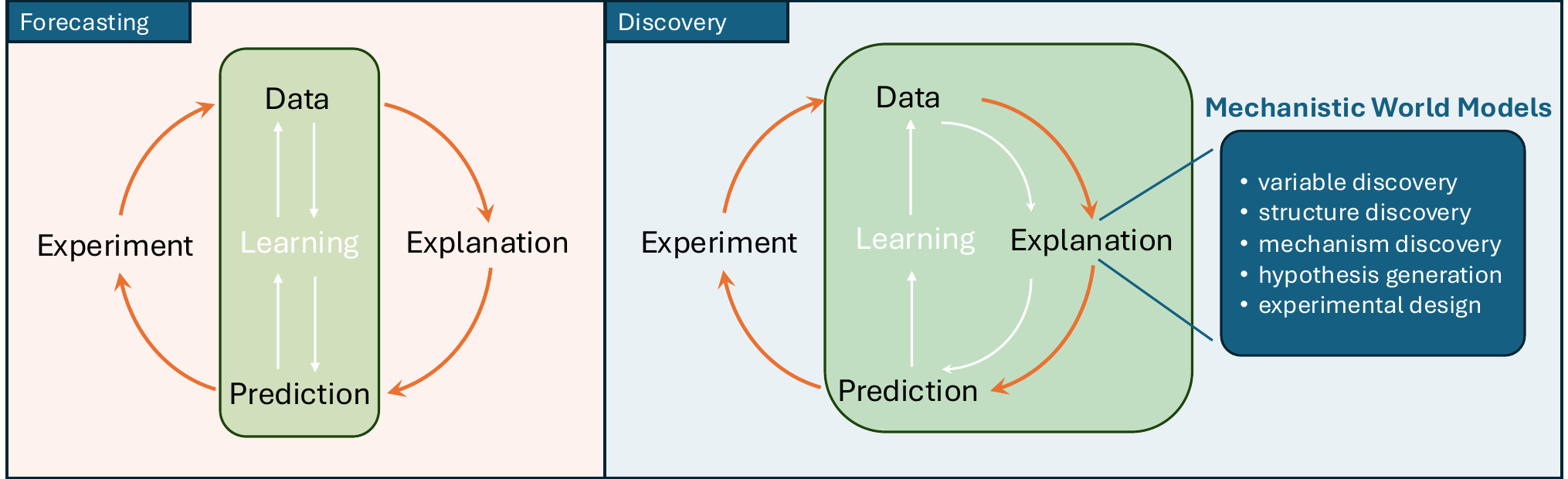}
    \caption{Mechanistic World Models (MWMs) shift AI from forecasting to discovery. Whereas conventional machine learning organises knowledge around predictive mappings, MWMs organise it around reusable mechanisms as the central computational abstraction linking prediction and experimentation. This enables capabilities including variable, mechanism and structure discovery, explanatory hypothesis generation and experimental design, extending machine learning beyond accurate forecasting towards scientific understanding.}
    \label{fig:mwm}
\end{figure}

The remainder of this paper develops this argument. We first revisit the philosophy of scientific discovery and explanation to derive the properties that a learned model must possess in order to support scientific discovery. We then examine how existing research directions in machine learning capture complementary aspects of these principles while falling short of a unified framework. Building on these foundations, we introduce the capabilities, design principles and anatomy of Mechanistic World Models (MWMs). By organising knowledge around reusable mechanisms, MWMs aim to bridge the gap between prediction and explanation while providing a computational substrate for scientific discovery, as illustrated in Figure~\ref{fig:mwm}. Finally, we discuss the capabilities afforded by this paradigm, including improved interpretability, compositional generalisation, targeted adaptation and scientific insight, and examine progress towards learning such systems together with the key challenges that remain.
\section{Scientific Discovery: A Philosophy of Science Perspective}\label{sec:pos}
Philosophers since Aristotle have argued that scientific inquiry begins with wonder at observed phenomena, but culminates only when we grasp their underlying causes (e.g.~\cite{AristotlePhysicsHardie, AristotlePosteriorAnalyticsMure, Losee2001HistoricalIntro}). The human scientific endeavour is fundamentally characterised by a transition from observing phenomena to uncovering the hidden structural principles that explain them. To understand the current limitations of machine learning in this area, we must first contrast mere prediction with genuine insight and highlight in particular the role that \emph{explanation} plays in generating reusable theories of how the world works. Inspired by the discourse in the Philosophy of Science about the nature of \emph{good} scientific  explanations, we will then extend our discussion to the nature of laws and broaden it to the concept of \emph{mechanisms}.

\subsection{Prediction vs Insight}
The distinction between \emph{description} and \emph{explanation} is a well-established cornerstone in the philosophy of Science (e.g.~\cite{HempelOppenheim1948,Salmon1990,Woodward2002Explanation}) and has equally been long recognised in machine learning (e.g.~\cite{shmueli2010explain}). A description details \emph{how} a phenomenon unfolds within a specific, observed regime; an explanation reveals \emph{why} it must unfold in that particular way. This division cleanly maps onto the epistemic gap between prediction and insight. While contemporary predictive models excel at capturing correlations to forecast complex protein structures or subsequent environmental states and observations, they remain fundamentally descriptive: even when equipped with domain-specific priors, they are not designed to autonomously discover the reusable structural invariants, such as physical variables and mechanisms, that underpin scientific understanding.

In contrast, scientific discovery depends entirely on our ability to move past prediction alone to generate explanatory insight. As highlighted in both cognitive and philosophical accounts of inquiry, \emph{the primary function of an explanation is not to merely track statistical regularities, but to expose the invariant structures of a system} (e.g.~\cite{Lombrozo2006, Woodward2004}). While a prediction satisfies the immediate \emph{how} or \emph{what}, an explanation addresses the fundamental question of \emph{why} a system responds to perturbation in a specific manner. 
This implies that scientific insight hinges not merely on the unconstrained ability to predict, but on the capacity to do so on the basis of a stable set of discovered regularities or abstractions. It is these invariant abstractions that fundamentally constitute insight (e.g.~\cite{HempelOppenheim1948, Salmon1990}). 
\emph{Prediction and explanation, therefore, are intimately linked}: better predictions should be achieved through abstractions lending themselves to ``good'' explanations. Clearly, the question of what constitutes a good explanation is pivotal and we shall return to it in the next subsection. Here, however, we highlight that machine learning traditionally has divorced prediction from explanation~\cite{breiman2001statistical, shmueli2010explain}. Driven by the significant recent successes of purely predictive models via scaling, a path often justified via the Bitter Lesson~\cite{sutton2019bitter}, the field has overwhelmingly prioritised purely predictive models over structural priors and treated explanation and interpretation largely as an afterthought through paradigms such as mechanistic interpretability~\cite{elhage2021mathematical, huben2024sparse}. 
Yet, there is a growing recognition that prediction and explanation should be explicitly linked to achieve robust, interpretable and scientifically useful learning while avoiding superficial shortcuts~\cite{herasimchyk2025prediction, hofman2021integrating}. We share this view. The remaining question is what properties a learned representation must possess in order to support explanation rather than prediction alone.

\subsection{The Anatomy of a Good Scientific Explanation}

The criteria for what constitutes a good explanation have evolved considerably over the past century. Early accounts emphasised logical deduction, later work shifted towards statistical and causal relationships, while more recent perspectives increasingly focus on uncovering the stable structures that generate observed phenomena. Rather than providing a comprehensive historical survey, we highlight the developments most relevant to the central thesis of this paper. Importantly, these accounts should not be viewed as successive replacements culminating in a single accepted theory of explanation. Rather, they emphasise complementary aspects of scientific explanation that together illuminate the properties explanatory models should possess. Readers interested in a broader treatment are referred to the historical review by Woodward~\cite{Woodward2002Explanation} and the comprehensive survey by Salmon~\cite{Salmon1990}.

The deductive-nomological (DN) model of Hempel and Oppenheim~\cite{HempelOppenheim1948} viewed explanation as the logical deduction of observations from general laws and initial conditions. Although enormously influential, the DN account was later criticised for conflating logical derivability with explanatory direction, as illustrated by the classic flagpole-and-shadow example~\cite{Salmon1990,Woodward2002Explanation}, where, given the angle of the sun, the height of a flagpole and the length of its shadow can each be logically deduced from the other, yet only the former \emph{explains} the latter. Subsequent statistical, causal and mechanistic accounts shifted attention from logical deduction towards identifying the factors genuinely responsible for an outcome, recognising that purely statistical relevance alone (for example the correlation between a falling barometer and an approaching storm) is insufficient for causal explanation~\cite{Salmon1997}. Mechanistic theories further shifted attention towards revealing the organised entities and activities that produce a phenomenon~\cite{Machamer_Darden_Craver_2000}. In parallel, unificationist theories proposed that the strongest explanations are those capable of accounting for many seemingly distinct phenomena using a small set of underlying principles~\cite{kitcher1989explanatory,Salmon1990}. Although no consensus has emerged on a single account of scientific explanation, there is increasing agreement that good explanations are characterised by their ability to identify stable, reusable structure. In particular, explanations are generally valued for combining parsimony, i.e. capturing phenomena using as few underlying principles as possible, with broad explanatory scope, i.e. accounting for many observations using those same principles. 

\subsection{Explanation as Information Organisation}
Taken together, these perspectives suggest that explanation can be viewed as a problem of \emph{information organisation}: finding the smallest collection of reusable mechanisms that explains the largest collection of observations.
Rather than treating explanations as isolated descriptions of individual phenomena, they should be considered as an organised hierarchy of abstractions that compress recurring patterns into reusable computational units. This view is closely aligned with long-standing ideas in cognitive science and systems theory. Marr~\cite{marr1982vision} argued that understanding an intelligent system requires identifying the appropriate levels of representation and abstraction, and further proposed a \emph{principle of modular design}: large computations should be decomposed into modules that are as nearly independent of one another as the task allows, since this is what makes a system tractable to debug, improve, or evolve. 
Likewise, Simon~\cite{simon1962architecture} observed that complex systems are often \emph{nearly decomposable}, meaning that they can be understood as collections of relatively independent subsystems whose interactions are substantially weaker than those within each subsystem. He further showed that this property, together with the fact that such systems are typically built from only a small number of recurring elementary subsystem types, is precisely what permits a complex structure to be given an economical, non-redundant description rather than one as cumbersome as the structure itself. Overall, this suggests that effective explanations arise not merely from accumulating information, but from organising it into reusable structures at the appropriate level of abstraction. 

Viewed through this lens, scientific discovery becomes a continual process of identifying recurring regularities and encapsulating them into mechanisms that can be reused across many observations. New observations are explained not by learning entirely new models, but by composing, refining or extending an existing library of mechanisms. As explanations become increasingly reusable, the amount of additional information required to explain new phenomena decreases while their explanatory scope increases. A discovery-oriented learning system should therefore be organised around representing the reusable mechanisms that generate observations rather than the observations themselves. This view unifies prediction and explanation: prediction concerns using an existing organisation of mechanisms to infer future observations, whereas explanation concerns discovering and improving that organisation itself. Section~\ref{sec:related_ml} examines how these principles relate to existing machine learning paradigms before introducing an architecture explicitly designed around them.

\subsection{Laws, Variables \& Mechanisms}
\label{sec:laws,variables,mechanisms}
Scientific explanations have traditionally been formulated in terms of laws, particularly in the physical sciences. More recent mechanistic accounts, however, argue that many phenomena, especially in biology, neuroscience and complex systems, are more naturally explained by identifying the mechanisms that generate them rather than by appealing solely to universal laws~\cite{Machamer_Darden_Craver_2000}. This shift reflects a broader view of scientific explanation: rather than seeking isolated predictive relationships, science seeks reusable explanatory structures that account for observations across increasingly diverse settings. 

We adopt mechanisms as the fundamental explanatory abstraction throughout this paper. In our view, mechanisms provide a unifying representation that encompasses scientific laws in domains where they exist (even if they are not yet known) while extending to other phenomena in domains where no general law may exist. Scientific laws are therefore not treated as distinct explanatory objects, but as mechanisms possessing exceptionally broad explanatory scope. Under this interpretation, learning mechanisms and discovering scientific laws become two instances of the same underlying problem: identifying reusable structures that explain observations. Mechanisms, therefore, provide a versatile computational abstraction for learning systems. Unlike individual predictive mappings, they are inherently local, reusable and composable, allowing increasingly complex explanations to be constructed from a relatively small library of explanatory building blocks.

Following~\cite{Machamer_Darden_Craver_2000}, mechanisms comprise organised \emph{entities} and \emph{activities} that produce regular changes. For the purposes of this paper, we therefore conceptually\footnote{We will formally define a mechanism for use in a Mechanistic World Model in Section~\ref{sec:mwm}.} define a mechanism as
\begin{equation}
    m \triangleq (\mathcal{V}, f), 
\end{equation}
where $\mathcal{V}$ corresponds to the entities (and their properties), while $f$ corresponds to the activities through which those entities evolve. A mechanism therefore couples entities with the transformation that acts upon them and may represent a physical law, a biochemical pathway, a neural circuit or any other reusable computational process that generates observations.

Mechanisms are consequently the fundamental building blocks of both prediction and explanation. Prediction is achieved by executing mechanisms to infer future observations, while explanation is achieved by discovering, representing and composing the mechanisms responsible for generating those observations. This suggests a fundamentally different objective for machine learning: rather than organising internal representations around observations alone, learning systems intended for scientific discovery should organise them around the reusable mechanisms that faithfully generate those observations. The following section examines the extent to which existing machine learning paradigms already embody this principle.
\section{From Philosophy of Science to Machine Learning} \label{sec:related_ml}
Thus far, we have argued from the perspective of the Philosophy of Science that scientific discovery depends on organising knowledge around reusable explanatory mechanisms rather than predictive mappings alone. This raises a second question: to what extent have these principles already begun to emerge within machine learning? Although the most visible successes of deep learning remain predominantly predictive, a substantial body of work across several active research communities has independently begun to emphasise many of the same ideas, including modularity, invariance, causal structure, and parsimonious representations. Viewed through the lens developed in the previous section, these fields can be understood as converging on complementary aspects of the same underlying problem: how should knowledge be organised to support explanation rather than prediction alone? In this section, we examine four representative directions, mechanistic interpretability, causal representation learning, equation discovery, and mixture of expert architectures and draw out the lessons each offers towards data-driven scientific discovery.

\subsection{Mechanistic Interpretability}
The abundant successes of predictive foundation models invite a promising route towards scientific discovery: one might develop methods to understand how models make predictions in the hope that the internal computations implemented by predictive models reflect the underlying mechanisms of the world. 
The field of mechanistic interpretability~\citep{elhage2021mathematical}, originally developed in the context of AI safety and alignment of large models, reverse-engineers how information is processed within a model. 
Methodological advances such as sparse autoencoders (SAEs)~\citep{huben2024sparse} and cross-layer transcoders (CLTs)~\citep{dunefsky2024transcoders, ameisen2025circuit} provide unsupervised methods for extracting interpretable features and abstractions within models' activation patterns.
A nascent and growing body of work now turns these tools onto scientific foundation models~\citep[e.g.][]{simon2025interplm, Gujral2025sparse, brixi2026genome}, raising the tantalising prospect of recovering features that correspond to scientific concepts without supervision.
 
Several representative cases illustrate what these methods have delivered and, more instructively, where they have fallen short. 
Recent works on protein language models have demonstrated the potential of mechanistic interpretability techniques for discovering biological concepts. Applied to models such as ESM-2~\citep{lin2023evolutionary}, SAEs recover latent features that can be interpreted as well-established concepts such as binding sites and structural motifs~\citep{simon2025interplm, Gujral2025sparse}. 
Yet, signs of success are far from universal.
Applications of the same toolset to single-cell foundation models such as Geneformer~\citep{theodoris2023transfer} and scGPT~\citep{cui2024scgpt} tell a more cautionary tale. A systematic investigation on attention patterns reveals computation circuits that reflect statistical co-expression rather than the underlying causal pathways of gene regulation~\citep{kendiukhov2026systematic}.
Another study~\citep{ahlmann2025deep} provides further evidence on a behavioural level that these models do not currently capture the causal mechanisms of gene regulation, showing that foundation models do not yet outperform simpler baselines in perturbation prediction tasks. 
The starkest example of the prediction-explanation gap comes from models that learn physics. Vafa et al.~\citep{vafa2025what} train world models on planetary orbits and investigate what the models have learned via probing, showing that models consistently fail to internalise Newtonian mechanics despite being able to accurately predict planet trajectories.

Taken together, this track record of mixed successes reveals a fundamental limitation of the model interpretation approach for scientific discovery: interpretability research is about interpreting the \emph{model}, not the \emph{world}. 
Crucially, mechanistic interpretability and other post-hoc analysis methods treat training and explanation as separate processes, leaving the training process unconstrained with respect to knowledge organisation.
Existing studies, such as the orbital mechanics investigation, provide evidence that the bottleneck is not that models are hard to interpret, but that models themselves often fail to internalise genuine mechanisms despite their prediction capability.  Viewed through the lens of Section~\ref{sec:pos}, the lesson is therefore that explanatory organisation cannot simply be extracted after training. Instead, it must be encouraged during the learning process itself.

\subsection{Causality and Representation Learning}
Mechanistic interpretability demonstrates that purely predictive objectives do not reliably recover such structures. Causality provides perhaps the most influential mathematical framework for formalising precisely this distinction between statistical prediction and underlying structure~\cite{scholkopf2022causality, pearl2009causality}. Here, causal discovery provides principled methods for inferring causal relationships from observational and interventional data. However, in their original form, these methods operate over a fixed and often pre-defined set of causal variables. In scientific settings, this is the crux of the difficulty: observations often come in the form of images, sequences, and other raw measurements, in which discovering the appropriate abstractions and variables is itself a central part of scientific discovery. This is precisely the challenge addressed by representation learning.

Disentangled representation learning from observation alone, however, is provably impossible without the right inductive bias, as there are infinitely many representations that fit the observations equally well~\citep{locatello2019challenging}.
Causal representation learning (CRL)~\citep{scholkopf2021towards} proposes to break this symmetry by imposing causal structures, seeking to recover latent variables that are consistent with key causal principles, such as the independent causal mechanisms principle and the sparse mechanism shift hypothesis.
Within these settings, CRL research has produced a substantial body of identifiability results that characterise how different conditions can lead to the recovery of causal variables~\citep{lachapelle2022disentanglement, kugelgen2023nonparametric, lippe2022citris}.
This theoretical rigour, however, comes at the cost of flexibility.
CRL approaches are typically confined to settings that fit the theoretical formalism they subscribe to, which limits their empirical success largely to synthetic benchmarks. 
As such, scaling the theoretical guarantees of CRL to open-ended scientific regimes, where unconstrained large models excel at prediction, remains elusive.

The enduring lesson from CRL, we argue, is not necessarily any one particular theoretical construction, but a general way of thinking about representations: desirable variables are those governed by mechanisms with the right structure. Stripping away the formal setup of specific instantiations, the core principles of causality express what this structure should look like. Sparse mechanism shift \citep{scholkopf2021towards} asks that most mechanisms remain unchanged as conditions change; the independent causal mechanism principle \citep{ParRojKilSch17} posits that mechanisms can be locally intervened upon and recombined without affecting one another; invariance demands that the same mechanisms remain valid across domains \cite{PetJanSch17,Huangetal20}. These principles closely mirror the characterisation of good explanations developed in Section~\ref{sec:pos}: explanatory mechanisms should be modular, stable and broadly \emph{reusable} across different settings. In this light, CRL offers an important route towards abstraction discovery in learning paradigms aimed at scientific discovery by suggesting how learning objectives can favour variables whose evolution is governed by reusable mechanisms.
 
\subsection{Equation Discovery}
The philosophical discussion also highlighted another defining characteristic of good explanations: they compress many observations into a small number of reusable principles. Equation discovery provides perhaps the clearest computational realisation of this idea within machine learning. Across much of the physical sciences, explanatory theories take the form of compact mathematical expressions governing system dynamics. Accordingly, equation discovery~\citep[e.g.][]{schmidt2009distilling, brunton2016dicovering} pursues the ideal of structured explanation by directly searching for closed-form equations that explain observations rather than relying on black-box function approximators. Indeed, success is typically evaluated not only by predictive accuracy but by symbol recovery itself~\citep{cava2021contemporary}.  Yet, symbolic regression faces an underdetermination challenge analogous to that of representation learning: many distinct equations can fit the same data. 
Here, we examine a design principle that has enabled the field to confront this challenge: \emph{parsimony}.

The principle of parsimony breaks the symmetry amongst candidate explanations by preferring functions that are built with fewer and simpler components. 
This notion is instantiated in different forms across the field. 
SINDy~\citep{brunton2016dicovering} casts parsimony as sparse regression over a predefined library of candidate terms, ensuring that only a small number of active terms participate in the explanation.
MDLformer~\citep{yu2025symbolic} expresses parsimony via an algorithmic-theoretic route, invoking the minimum description length principle~\citep{grunwald2007minimum} to directly penalise the complexity of the learned equation.
In other neural network augmented hybrid approaches, such as AI Feynman~\citep{udrescu2020ai} and symbolic distillation~\citep{cranmer2020discovering}, complexity regularisation on candidate equations also plays an enabling role in recovering the underlying equations for physics systems. 
Together, this convergence towards simplicity across diverse approaches underscores the indispensable role of parsimony in discovery. 

Yet, successes in equation discovery methods are primarily clustered around rediscovering isolated physical systems with known governing equations, the regimes for which these methods are built.
Extending the symbolic framework to broader, larger-scale, open-ended settings runs into two obstacles. 
Firstly, akin to the challenge with classical causal discovery, learning symbolic expressions requires access to the variables at the right level of abstraction, the discovery of which remains an open challenge. 
Secondly, and more fundamentally, equation discovery is about retrieving a single universal governing law that models a particular phenomenon, rather than capturing composable explanatory mechanisms (see discussion in Section~\ref{sec:laws,variables,mechanisms}).

That low-complexity models generalise better is a central idea dating back to statistical learning theory. Equation discovery, however, illustrates a deeper role for parsimony. Across diverse algorithms and application domains, simplicity consistently determines not only predictive performance but whether meaningful explanatory structure emerges at all. In light of Section~\ref{sec:pos}, parsimony is therefore more than a statistical regulariser, it is a requirement: it governs how explanatory information is organised within a model. While autonomous scientific discovery need not ultimately employ symbolic representations, equation discovery provides compelling evidence that explanatory learning requires explicit pressure towards compact, computational structure.

\subsection{Information Organisation via Mixtures of Experts}
Among modern large-scale architectures, Mixture of Experts (MoE) models (e.g.~\cite{shazeer2017moe,fedus2022switch,ParRojKilSch17}) arguably provide the clearest empirical demonstration that how information is organised within a model fundamentally affects its capability. Rather than distributing computation uniformly across all parameters, MoEs partition computation across a collection of specialised subnetworks, with a learned routing function selecting only a small subset of experts for each input. This sparse computation enables models to scale to unprecedented parameter counts while substantially reducing computational cost. More importantly, MoEs often exhibit functional specialisation, with different experts becoming responsible for different classes of inputs or capabilities. Organising knowledge into relatively independent computational units has therefore been demonstrated to be an effective inductive bias for large-scale learning. Moreover, by localising computation and gradients, MoE architectures have the potential to facilitate continual acquisition of new competencies, a capability recently demonstrated in the context of skill learning~\citep{wang2024sparse}. In this sense, they offer compelling evidence that modular organisation is a promising direction for scalable learning. However, the organising principle underlying MoEs remains fundamentally predictive. Experts are learned because they improve optimisation of the predictive objective, not because they correspond to reusable mechanisms governing the underlying system. As such, while these decompositions improve efficiency, they need not align with the abstractions that underpin scientific explanation. 

From the perspective of mechanistic explanations, the central question is therefore not whether models should be modular, but what the modules should represent. Rather than partitioning computation according to predictive competence, discovery requires the organisation of representations around reusable mechanisms that remain valid across observations, interventions, and domains. Such a structure localises both computation and learning: new evidence updates only those mechanisms implicated in the observed phenomenon, while leaving unrelated mechanisms unchanged. This provides a principled route towards mitigating catastrophic forgetting, improving interpretability, and enabling the compositional reuse of scientific knowledge.

\subsection{Machine Learning for Scientific Discovery}
Although these research directions differ substantially in their assumptions, objectives and methodologies, they exhibit a striking convergence. Together, they independently reinforce the central thesis developed in the previous section: scientific discovery depends fundamentally on how knowledge is organised. While each of these research directions captures an important aspect of explanatory learning, they do so largely in isolation. We argue that learning systems for scientific discovery require integrating these complementary ingredients into a common substrate of reusable mechanisms.
\section{Mechanistic World Models} \label{sec:mwm}
Sections~\ref{sec:pos} and~\ref{sec:related_ml} approach the problem of scientific discovery from complementary perspectives. Philosophy of Science identifies the properties explanatory models must possess to enable discovery, while a review of related concepts in machine learning reveals a growing convergence towards many of these same ideas through modularity, causality, parsimony and structured representations. In this section we synthesise these ideas into a coherent design paradigm: \emph{Mechanistic World Models} (MWMs). We build upon world models because they explicitly represent system dynamics under observation and intervention. Furthermore, prediction remains essential: as argued in the previous sections, a model cannot explain a system's evolution without also predicting how it will behave. MWMs therefore extend, rather than replace, conventional world models by shifting the representational objective from observations to mechanisms. 
Mechanistic World Models can similarly be viewed as mechanism-centric latent state-space models. Like conventional state-space models, they encode observations into a latent state from which the system's future evolution is predicted. Within an MWM, this state comprises both (latent) variables describing the current state of the system and latent parameters governing the behaviour of its mechanisms. The key distinction is that the transition dynamics are no longer represented as a monolithic function, but are instead organised into reusable mechanisms and the structures that instantiate and compose them. 
Crucially, variables and mechanisms are not discovered independently. Variables emerge because they admit compact, reusable mechanisms, while mechanisms emerge because they operate over variables that support reusable explanations.

We first highlight the benefits afforded by discovery-oriented learning systems and derive the capabilities they must possess. We then identify the design principles and inductive pressures required for these capabilities to emerge, before finally describing the anatomy of a Mechanistic World Model.

Mechanism-centric organisation fundamentally changes not only what a world model represents, but also the properties it affords. By organising knowledge around reusable mechanisms rather than predictive mappings, Mechanistic World Models promise \emph{interpretability by design}, \emph{compositional generalisation}, and more effective \emph{continual learning}, through localised adaptation. Most importantly, they move beyond prediction towards \emph{explanatory knowledge}: rather than merely forecasting future observations, they aim to reveal the mechanisms that generate them. Realising these benefits, however, requires capabilities that extend substantially beyond conventional world modelling.
These capabilities fall into two broad groups: \emph{knowledge organisation} and \emph{active inquiry}. The former concerns discovering reusable explanatory abstractions from observations, while the latter concerns using these abstractions to drive the scientific process itself.

\paragraph{Knowledge Organisation}

Given observation streams from different domains and systems, anMWM must organise its knowledge around reusable explanatory abstractions. This requires jointly discovering variables, mechanisms and the structures that relate them. \textbf{Variable discovery} requires identifying quantities that represent the state of a system at the level of abstraction where explanation becomes possible. Equally, \textbf{mechanism discovery} requires learning versatile, reusable transformations that capture how variables evolve and influence one another. Finally, variables and mechanisms must be related through \textbf{structure discovery}, identifying how they are bound together to explain a particular system. It is this binding structure that enables mechanisms to be reused and recomposed across related settings rather than relearned independently for each new environment.

\paragraph{Active Inquiry}

Knowledge organisation alone, however, is insufficient for scientific discovery. Instead, it requires an active process of proposing explanations and seeking the evidence needed to validate or refine them. An MWM must therefore support \textbf{explanatory hypothesis generation}, proposing candidate variables, mechanisms and structures that extend beyond previously observed systems through recombination of its existing knowledge. To close this discovery loop, it must also support \textbf{experimental design}, using these hypotheses to identify the observations and interventions expected to be maximally informative for distinguishing between competing explanations and refining the explanatory model.

Together, these capabilities give rise to the benefits outlined above. Explicit variables, mechanisms and structures make learned knowledge interpretable by construction rather than through post-hoc analysis. Reusable mechanisms enable compositional generalisation across related systems. Because mechanisms act on explicit variables, they also support reasoning about interventions by localising the consequences of changes to the relevant mechanisms or variables. This encapsulation also localises learning, allowing new evidence to refine only the mechanisms implicated by an observation, thereby mitigating catastrophic forgetting. Most importantly, organising knowledge around mechanisms transforms world models from systems that merely predict how the world will evolve into systems that explain why it evolves in that way, closing the gap between forecasting and scientific understanding.

\subsection{Design Principles \& Inductive Pressures}
\label{sec:design_principles}
Whether the capabilities outlined above will ultimately emerge purely by scaling predictive models in capacity, data and compute remains an open question. However, our thesis is that they should instead be encouraged to emerge explicitly by designing learning systems whose architectures and learning objectives are aligned with the principles of scientific discovery developed in Section~\ref{sec:pos}. In particular, the Philosophy of Science identifies the properties that make explanations scientifically useful, while the machine learning literature demonstrates that many of these properties can be encouraged through appropriate inductive biases. Together, these perspectives motivate four core design principles for Mechanistic World Models, as well as two principal inductive pressures that encourage their emergence.

\paragraph{Design Principles} 
\begin{enumerate}
    \item \textbf{Joint predictive \& explanatory modelling.} Prediction provides the supervisory signal, while explanation determines how knowledge is organised. Explanatory structure such as variables, mechanisms and their relationships should emerge jointly during optimisation rather than through post-hoc analysis. This principle follows directly from our central thesis that prediction and explanation should not be divorced.

    \item \textbf{Mechanisms as the fundamental unit of representation.} In line with the mechanistic view of scientific explanation, knowledge should be organised around reusable mechanisms acting on explicit variables rather than arbitrary latent representations. Mechanisms, defined as variable-transformation pairs, become the primary computational abstraction through which knowledge is represented, composed and refined.
    Structural causal models illustrate one probabilistic instantiation of this viewpoint~\citep{pearl2009causality,PetJanSch17}, in which mechanisms correspond to structural assignments over a fixed causal graph. MWMs instead organise knowledge around reusable mechanisms that can be discovered, instantiated and composed across different systems. Alternative instantiations have been proposed where functions are programs run on Turing machines, with bit strings playing the role of exogenous variables (e.g.~\citep{JanSch10}).

    \item \textbf{Mechanism reuse.} Scientific understanding derives much of its power from explaining diverse phenomena through a relatively small number of shared mechanisms. Rather than repeatedly learning new predictive mappings, MWMs should preferentially reuse existing mechanisms across systems, environments and domains, refining existing mechanisms and introducing new ones only where necessary. This requires mechanisms to remain approximately invariant and independently reusable across different contexts, consistent with the Independent Causal Mechanisms principle and the Sparse Mechanism Shift hypothesis discussed in Section~\ref{sec:related_ml}. 
    
    \item \textbf{Hierarchical organisation of mechanisms.} As argued by Marr and Simon (Section~\ref{sec:pos}), complex systems can be understood as hierarchies of relatively independent abstractions. Mechanisms should therefore be organised hierarchically, allowing simple mechanisms to compose into increasingly rich explanatory structures while maintaining modularity and interpretability.
\end{enumerate}

While these principles define what a Mechanistic World Model should represent, they do not determine how such representations emerge. This is the role of inductive pressures.

\paragraph{Inductive Pressures}

The Philosophy of Science and the machine learning literature reviewed in the previous sections repeatedly converge on two overarching inductive pressures: \emph{parsimony} and \emph{compositionality}. Neither is new. Parsimony has long underpinned, amongst others, scientific explanation, statistical learning and equation discovery, while compositionality has played a central role in modular architectures, causal models and structured representation learning. The key distinction is that these principles are typically employed to improve optimisation, regularisation or computational efficiency. In Mechanistic World Models, they instead act as pressures for \emph{organising explanatory knowledge}. Together, these pressures encourage the emergence of compact, reusable variables and mechanisms that can be recomposed across diverse systems.

\begin{enumerate}
    \item \textbf{Parsimony.} As discussed in Section~\ref{sec:pos}, good scientific explanations account for many diverse observations using the smallest possible collection of explanatory principles. Accordingly, MWMs favour variables, mechanisms and structures that jointly minimise representational complexity while maximising explanatory scope. Parsimony is therefore applied not merely to functions or parameters, but to the organisation of explanatory knowledge itself.
    
    \item \textbf{Compositionality.} Scientific knowledge accumulates by composing existing explanations into increasingly sophisticated theories rather than replacing them wholesale. Likewise, MWMs favour explanations that can be constructed through the recombination of existing mechanisms. This pressure encourages reusable libraries of mechanisms and supports compositional generalisation to previously unseen systems. It also localises adaptation during continual learning by favouring the modification of existing explanatory components over learning entirely new ones.
\end{enumerate}
Together, these design principles and inductive pressures provide the conceptual blueprint for Mechanistic World Models. Next, we describe the resulting anatomy of such systems.

\subsection{The Anatomy of a Mechanistic World Model}
\label{sec:anatomy}
\begin{figure}[t]
    \centering
    \includegraphics[width=0.99\linewidth]{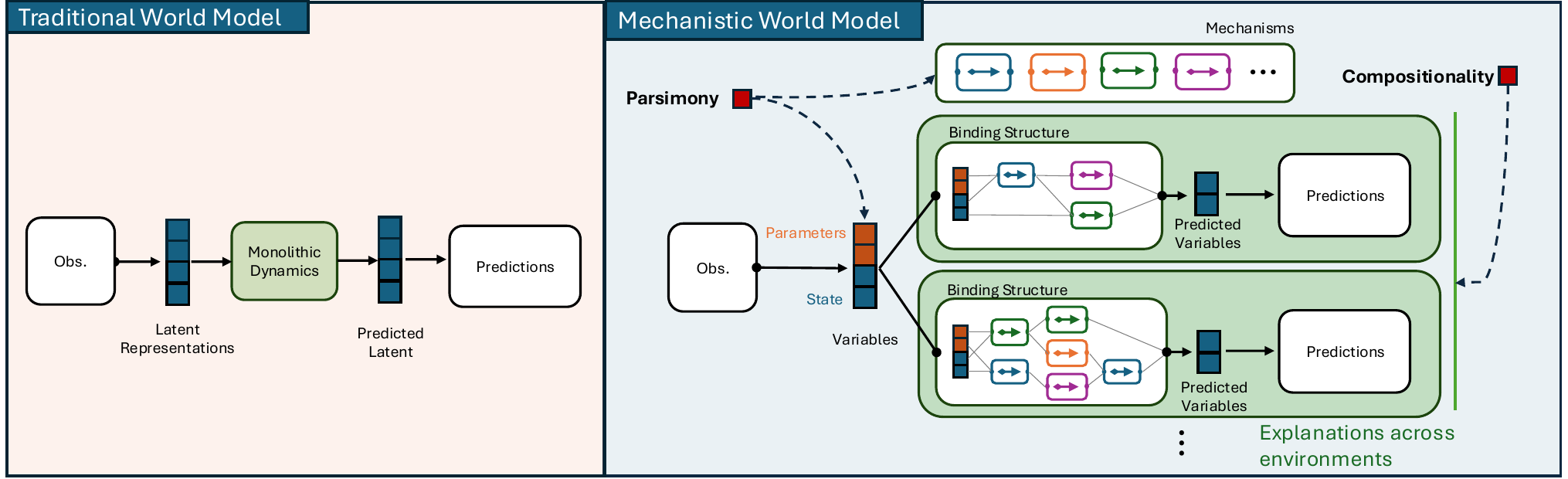}
    \caption{Conceptual anatomy of a Mechanistic World Model. Observations are encoded into latent variables representing the state of a particular system. Reusable mechanisms, each comprising variable roles and a transformation, are instantiated by binding these roles to the learned variables. Different systems are explained through different structures that instantiate and compose a shared library of reusable mechanisms. Parsimony favours compact variables, simple mechanisms and sparse explanatory structures, while compositionality encourages the reuse and recombination of mechanisms across different systems through different explanatory structures.
    }
    \label{fig:WMW_anatomy}
\end{figure}

The capabilities and design principles developed above lead to a simple conceptual anatomy for Mechanistic World Models. Rather than organising knowledge around predictive mappings, MWMs organise knowledge around three fundamental components: \emph{variables}, \emph{mechanisms}, and the \emph{structure} that instantiates and composes them into explanations of individual systems (Fig.~\ref{fig:WMW_anatomy}).

\paragraph{Variables}
Variables describe the (latent) state of a system together with latent parameters governing its dynamics. Learned jointly with mechanisms, they constitute the state space over which prediction and explanation are performed. Variables are therefore not discovered in isolation, but emerge together with the mechanisms they support: useful variables are precisely those that admit compact, reusable mechanisms. We first define a set of semantic variable types,
\begin{equation}
\mathcal T=\{\tau_1,\tau_2,\ldots\},
\end{equation}
which capture the \emph{functional role} of variables. These types are learned jointly with the variables, mechanisms and structures themselves and represent emergent functional roles within the model, analogous to quantities such as position, velocity or mass, rather than pre-specified semantic labels. A Mechanistic World Model then represents a particular system through a collection of latent variables
\begin{equation}
Z=\{z_1,\ldots,z_n\},
\qquad
T:Z\rightarrow\mathcal T,
\end{equation}
where the typing function $T$ assigns each latent variable a semantic type.

\paragraph{Mechanisms}
Following the \emph{conceptual} definition introduced in Section~\ref{sec:pos}, each mechanism specifies how variables of particular semantic types interact. We now formally define a mechanism as
\begin{equation}
m\triangleq(\Sigma_m,f_m),
\end{equation}
where
\begin{equation}
\Sigma_m=
\big(
(\tau_1,\ldots,\tau_p),
(\sigma_1,\ldots,\sigma_q)
\big),
\qquad
\tau_i,\sigma_j\in\mathcal T,
\end{equation}
is the typed variable signature specifying the semantic roles of the input and output variables, and $f_m$ is the corresponding deterministic or probabilistic mapping from the input roles to the output roles.
Since the signature refers only to semantic variable types rather than particular latent variables, mechanisms are reusable across different systems and domains. Mechanisms therefore constitute the fundamental explanatory abstraction of an MWM, while their explicit encapsulation enables refinement, composition and reuse with minimal interference\footnote{When instantiated probabilistically, this formulation is closely related to structural causal models, in which mechanisms correspond to structural assignments. The key distinction is that MWMs treat mechanisms as reusable computational units that can be instantiated through different bindings across different systems, rather than as structural assignments attached to a fixed causal graph over predefined variables.}.

\paragraph{Structure}
Structure specifies how learned latent variables instantiate the variable roles of reusable mechanisms and how these instantiated mechanisms are composed to explain a particular system. A mechanism is instantiated through a binding
\begin{equation}
b:\Sigma_m\rightarrow Z,
\end{equation}
where, by slight abuse of notation, $\Sigma_m$ denotes the set of variable roles appearing in the mechanism signature such that $b$ assigns each variable role in the mechanism signature to a latent variable. Valid bindings satisfy the type constraint
\begin{equation}
T(b(\tau))=\tau,
\label{eq:type_matching}
\end{equation}
ensuring that each role is filled by a variable of matching semantic type. Different environments are therefore represented not through different mechanisms, but through different collections of bindings that instantiate and compose a shared mechanism library.

\vspace{0.5em}\noindent
Together, semantic types, latent variables, mechanisms and bindings constitute an explicit, interpretable explanation of how a system behaves. Formally, we therefore define a Mechanistic World Model as the tuple
\begin{equation}
\mathcal W\triangleq(\mathcal T,Z,\mathcal M,S),
\end{equation}
where $\mathcal T$ is the set of semantic variable types, $Z$ is the set of typed latent variables, $\mathcal M=\{(\Sigma_m,f_m)\}$ is a library of reusable mechanisms, and $S=\{b_i\}$ is the collection of bindings that instantiate these mechanisms such that they compose into an explanation of a particular system.

Through this formulation, the two inductive pressures introduced in Section~\ref{sec:design_principles} can be applied to specific parts of the anatomy: \emph{parsimony} constrains the variables and mechanisms to minimise representational complexity, while \emph{compositionality} acts on the binding structures across different domains and systems using the same set of building blocks. Given a collection of observed environments, $\mathcal{E}$, learning an MWM can be schematically expressed as
\begin{equation}
\mathcal{W}^* = 
\operatorname*{arg\,min}_{\mathcal{T},\, Z,\,  \mathcal{M},\, \{S_e\}_{e \in \mathcal{E}}}
\quad
\left[\Omega_p\big(\mathcal{T}, \mathcal{M}, Z\big)
\;+\; \Omega_c\big(\{S_e\}\big)
\;+\; \sum_{e \in \mathcal{E}} \mathcal{L}_{\mathrm{pred}}\big(\mathcal{T}, Z,\mathcal{M}, S_e\big)
\right],
\label{eq:mwm_objective}
\end{equation}
where $S_e$ is the environment-specific binding patterns, $\Omega_p$ penalises the complexity of the learned variables, mechanisms, and types, and $\Omega_c$ measures the extent to which mechanisms are composed and reused across the binding structures of different environments, and $\mathcal{L}_{pred}$ is a prediction loss. 
Together, these pressures encourage what a good explanation requires: a small library of reusable variables and mechanisms that can be composed to explain diverse phenomena across different environments. 

\section{Towards Mechanistic World Models}
The promise of Mechanistic World Models is not to automate the activities that scientists perform, but to automate insight extraction from observations. 
Since the emergence of large language models, a growing number of systems have been proposed under the banner of AI Scientists, Autonomous Discovery, and AI for Science (e.g.~\citep{lu2026towards,gottweis2026coscientist,ghareeb2026multiagent})\footnote{Earlier Robot Scientist systems similarly automate aspects of scientific discovery (e.g.~\citep{sparkes2010robot,gower2024airobotic}), but focus primarily on automating the scientific workflow rather than the underlying representation of explanatory knowledge.}. These systems have demonstrated impressive capabilities for experiment planning and literature synthesis. However, they largely inherit the representational substrate of today's predictive foundation models. While agentic workflows may automate aspects of the scientific process, they fundamentally do not address the central challenge identified in this paper: how explanatory knowledge itself should be represented and acquired.

Realising this vision, however, remains an open challenge. While no existing method instantiates a Mechanistic World Model in full, a growing body of work has made substantial progress on its constituent challenges.
Rather than surveying the broader literature, in this section we focus on representative world modelling approaches that explicitly address one or more of the three principal anatomical components introduced in Section~\ref{sec:anatomy}: variable, mechanism and structure discovery. Each of these components has seen significant progress recently, though largely in isolation. Table~\ref{tab:comparison} summarises how these works relate to the capabilities of an MWM.

\begin{table*}[t]
\centering
\small
\caption{Comparison of representative approaches against the proposed design principles for Mechanistic World Models. While existing methods each address important aspects of scientific discovery, they typically focus on individual capabilities such as variable, mechanism, and structure discovery, compositionality, and parsimony. We further distinguish approaches by whether they are built on architectural building blocks with a credible path to scale: methods whose mechanistic structure is expressed through standard, scalable components rather than through bespoke architectures or search procedures. Mechanistic World Models aim to unify these complementary directions within a single framework organised around reusable mechanisms as the fundamental computational abstraction.
\label{tab:comparison}}

\begin{tabular}{lcccccc}
\toprule
\textbf{Method} &
\rotatebox{90}{\makecell{Variable\\Discovery}} &
\rotatebox{90}{\makecell{Mechanism\\Learning}} &
\rotatebox{90}{\makecell{Structure\\Learning}} &
\rotatebox{90}{\makecell{Compositionality}} &
\rotatebox{90}{\makecell{Parsimony}} &
\rotatebox{90}{\makecell{Scalable\\Architecture}} \\
\midrule

VCD~\citep{lei2023variational}
& \cmark & --     & \cmark & \cmark & \cmark & -- \\
COMET~\citep{lei2024compete}
& --     & \cmark & \cmark & \cmark & --     & -- \\
SPARTAN~\citep{lei2025spartan}
& --     & --     & \cmark & --     & \cmark & \cmark\\
DIDS~\citep{baumgartner2026disentangling}
& \cmark & --     & \cmark & --     & \cmark & -- \\
C-JEPA~\citep{nam2026causaljepa}
& --     & --     & \cmark & --     & --     & -- \\
RIM~\citep{goyal2021recurrent}
& --     & \cmark & \cmark & \cmark & --     & -- \\
C-SWM~\citep{Kipf2020Contrastive}
& \cmark & --     & \cmark & --     & --     & -- \\
DreamCoder~\citep{ellis2023dreamcoder}
& --   & \cmark & --     & --     & \cmark & -- \\
NEO~\citep{baek2026learning}
& --     & \cmark & --     & \cmark & \cmark & -- \\
PoE-World~\citep{piriyakulkij2025poeworld}
& --    & \cmark & --     & \cmark & -- & -- \\

\midrule
Mechanistic World Models & \cmark & \cmark & \cmark & \cmark & \cmark & \cmark \\

\bottomrule
\end{tabular}
\end{table*}

\subsection{Variable Discovery}
Discovering the variables to represent the latent state and the dynamical parameters from observations is a central capability of MWMs. Unlike classical causal discovery or equation discovery, which presuppose a fixed set of directly observable variables, an MWM must, from data, learn the right abstractions that admit mechanistic explanations. 

To this end, early works in object-centric representation learning, such as GENESIS~\citep{engelcke2021genesisv2} and SlotAttention~\citep{locatello2020object}, offer one potential route by developing architectural bias towards object-factored representation. These methods aim to decompose high-dimensional observations into discrete object slots without supervision and have served as the substrate for many downstream structure discovery approaches~\citep[e.g.][]{lei2025spartan, lei2024compete, nam2026causaljepa}.
However, object slots are not yet explanatory variables.
As argued throughout the paper, the abstractions that matter for discovery are those that admit compact and reusable downstream mechanisms. As such, while object slots may be a natural choice for some cases, whether they are the right units for reasoning, as opposed to dynamical properties, object parts, or other abstractions, needs to be determined jointly with the downstream dynamics.

This coupling motivates methods in which representations are shaped by the dynamics they support. C-SWM~\citep{Kipf2020Contrastive} shows that object-centric representation can emerge through joint training with a transition model that is constrained to represent relational message-passing structures.
More recently, causal representation learning methods have pushed this idea of dynamics-driven representation further by developing the theoretical foundations for when and how variable recovery can be made possible. VCD~\citep{lei2023variational} demonstrates that a causally structured dynamics model with sparsity regularisation can surface disentangled state variables from observation, while DIDS~\citep{baumgartner2026disentangling} uses a similar construction to establish the conditions under which dynamical parameters can be provably recovered.

Neither line completely resolves variable discovery for MWMs, but together these efforts provide some of the key ingredients. Object-centric representation learning develops scalable architectures that can explicitly and flexibly capture discrete entities from observations, while causal representation learning demonstrates, if only in restricted settings, how theoretical frameworks can ground variable discovery. Taken together, these approaches now outline a feasible path towards scalable and principled variable discovery for mechanistic world models. 

\subsection{Mechanism Discovery}
Given a set of variables, the MWM must discover the set of reusable mechanisms that govern how they evolve and interact in different environments. 
The central desideratum for these mechanisms is that they need to be recovered as distinct, encapsulated transformations that can be composed and independently refined across domains and settings. 

One direct instantiation of this idea comes from works that represent individual mechanisms as explicit programs. DreamCoder~\citep{ellis2023dreamcoder}, for example, uses neural-guided search to learn reusable programmatic primitives and maintains a growing library of programs that solves various computational tasks. More recently in the context of compositional world models, PoE-World~\citep{piriyakulkij2025poeworld} uses language models to synthesise small programs that are in turn composed as a product-of-expert world model.
These approaches illustrate the strengths of explicit encapsulation between mechanisms, namely, improved interpretability, data-efficiency, and modular updates. 
However, the limitations of this class of methods are clear: they require symbolic inputs with relatively simple dynamics, and have so far been confined to symbolic reasoning or game-like environments.

Another line of work pursues these properties within a learnable substrate. These approaches represent mechanisms as independently parameterised and learnable models, and investigate how distinct, specialised mechanisms can be made to emerge during training.
RIMs~\citep{goyal2021recurrent} serve as an early demonstration that the architectural bias of separately parameterised recurrent modules can develop functions that specialise in distinct mechanisms. More recently, NEO~\citep{baek2026learning} casts mechanism learning as a codebook learning problem, and uses the minimum description length principle to induce latent codes that correspond to composable dynamics primitives. 
These primitives, however, act sequentially on monolithic latent representations of observations rather than on individual variables. 
Closest to our MWM framework is COMET~\citep{lei2024compete}, which learns a library of modules that act on the object level via a competition-of-experts learning algorithm. 
By updating only the relevant part of the mechanism library and learning to compose, COMET leverages the encapsulated modules to adapt to different environments. 

These approaches showcase the feasibility and benefits of maintaining a library of reusable mechanisms across different tasks and domains. 
In particular, methods such as DreamCoder, NEO, and COMET offer potential instantiations of the parsimony and compositionality pressures: DreamCoder and NEO use minimum description length to encourage parsimony, while COMET uses a competition scheme to ensure compositional reuse.
While current methods remain proofs-of-concept that are difficult to scale, the design principles they embody, compositionality and parsimony, offer a foundation on which more scalable mechanistic models can be built. 

\subsection{Structure Discovery}
The remaining component is the structure that binds variables and mechanisms together to form explanations. This learnable structure needs to capture which mechanisms act on which variables, as well as how these mechanism–variable bindings are configured across systems.
Most progress in this space has come through the inference of interaction structure, i.e., recovering which entities influence which over a given set of variables. NRI~\citep{kipf2018neural} first cast this structure as a latent variable to be inferred from observation. More recent efforts in the context of structured world models have explored how such interaction structure can be captured by object-factored dynamics models. C-SWM~\citep{Kipf2020Contrastive} learns relational transitions between object slots, but does so with fully-connected GNNs that assume interactions among all entities rather than recovering which are actually present. C-JEPA~\citep{nam2026causaljepa} instead surfaces interaction structure implicitly, masking object slots so that the prediction for each entity becomes independent of irrelevant objects. SPARTAN~\citep{lei2025spartan} makes the interaction structure explicit by using sparsity-regularised attention to recover context-dependent interaction graphs. Crucially, the same sparse-attention construction has been shown to scale: in the context of mechanistic interpretability, it induces simpler internal circuit structures in models of up to seven billion parameters~\citep{draye2026intrinsically}, suggesting that sparse attention is a viable mechanism for recovering explanatory structure at scale.

However, recovering which variables interact is only part of the structure learning task.
A full explanatory structure must also bind specific, reusable mechanisms to the variables they act on, and reconfigure these bindings to explain varying systems. 
Here, COMET~\citep{lei2024compete} serves as a first step towards this direction. Its composition step uses dot-product attention to select, for each object, which mechanism from the library applies, and reconfigures these binding patterns across environments. 
This selective binding process can be interpreted as an early, implicit instance of the type-matching constraint concept in the MWM formulation (equ.~\ref{eq:type_matching}). COMET's bindings, however, remain limited to pairwise interactions and do not yet scale. Since binding is fundamentally a selection problem, the same sparse-attention construction that scales interaction structure in SPARTAN offers a plausible route to scaling mechanism binding as well.

\subsection{Challenges \& Opportunities}
The discovery of variables, mechanisms, and structures has each seen progress, as the preceding sections show.
Here, we foreground some of the outstanding challenges and opportunities that need to be addressed.

First, methodological advances almost always target individual facets of MWM learning in isolation, and to date, no existing solution fully integrates these components. 
This is not simply a matter of combining existing methods: putting together all of the components will require optimising over variables, mechanisms, and binding structures jointly while balancing the pressures of parsimony and compositionality. 
This integration also presents theoretical challenges around the identifiability of variables and mechanisms, including non-markovian systems and partial observability, which no existing theoretical framework can fully address. 

Secondly, \emph{active inquiry} capabilities, essential for realising the full potential of the mechanistic view of modelling, remain largely underexplored in the current literature.
The explicit knowledge organisation in MWMs promises to facilitate structured adaptation and information acquisition beyond passive learning. 
To reap the benefits of the encapsulated nature of mechanisms, however, requires machinery for mechanism management: deciding when to add a new mechanism; when to merge or refine existing ones; what information to seek in order to verify, refine or extend existing mechanisms; and how to reorganise the mechanism library as understanding deepens. In many domains, substantial mechanistic knowledge already exists, suggesting that MWMs should be able to initialise their mechanism library with existing knowledge rather than learning entirely from scratch. Determining how best to integrate prior mechanistic knowledge into a unified learning framework, while allowing it to be refined as new evidence becomes available, also remains an open question.

Finally, progress on any of these frontiers will require adequate \emph{evaluation} that current benchmarks do not yet provide. 
Assessing an MWM demands tasks in which the underlying variables, mechanisms, and structures are well understood, so that the model's recovered abstractions can be verified in a principled way rather than relying on prediction accuracy alone. 
Furthermore, such evaluation frameworks must also probe the capabilities that distinguish an MWM, testing whether a model is able to adapt to and learn from novel situations by efficiently reusing existing knowledge, and continuing to learn without overwriting what it already knows. 
Developing environments and metrics for such evaluation will be a crucial step in translating this conceptual vision into a tangible and measurable learning paradigm. 
\section{Conclusions}
Scientific progress depends not merely on predicting observations, but on uncovering the mechanisms that explain them. While modern machine learning has achieved remarkable success, its internal representations primarily serve forecasting rather than explanation. We argue that this distinction is fundamental: prediction is a prerequisite for scientific discovery, but it is not sufficient. Discovery requires learning systems that organise knowledge around reusable, explanatory abstractions.

Drawing on insights from both the Philosophy of Science and contemporary machine learning, we introduce \emph{Mechanistic World Models} as a conceptual framework towards this objective. Rather than treating representation learning, causality and interpretability as independent research directions, MWMs place mechanisms at the centre of representation, computation and learning. This perspective motivates the capabilities required for scientific discovery, including variable, mechanism and structure discovery, together with the design principles and inductive pressures that encourage such representations to emerge.

Although no existing system fully realises this vision, many of its constituent ingredients have emerged independently across diverse areas of machine learning. Advances in causal representation learning, modular architectures, equation discovery and mechanistic interpretability each illuminate important aspects of explanatory learning. Together, they provide many of the ingredients required for Mechanistic World Models, but have yet to unify them within a common framework organised around reusable mechanisms.

Much remains to be understood. Many of the central challenges identified in this paper remain open research problems. Equally, it is possible that sufficiently capable predictive models may eventually exhibit some of these properties through scale alone. Our thesis, however, is that scientific discovery is fundamentally a problem of knowledge organisation, and that architectures and learning objectives explicitly designed around reusable mechanisms provide a more direct route towards explanatory intelligence.

If artificial intelligence is to become a genuine engine of scientific discovery rather than merely remain a tool for prediction, it must move beyond forecasting to organising knowledge around reusable explanatory mechanisms. Mechanistic World Models provide a conceptual and computational foundation for that transition.






\bmhead{Acknowledgements}
This research was supported by an EPSRC Programme Grant (EP/V000748/1). The authors gratefully acknowledge the role of the ELLIS PhD Programme in bringing together the authors to lay the foundations for this work.

\bibliography{references}

\end{document}